# QKVShare: Quantized KV-Cache Handoff for Multi-Agent On-Device LLMs

Pratik Honavar

Tejpratap GVSL

pratikhonavar142857@gmail.com    tejpratap.gvsl@gmail.com

## Abstract

Multi-agent LLM systems on edge devices need to hand off latent context efficiently, but the practical choices today are expensive re-prefill or full-precision KV transfer. We study QKVShare, a framework for quantized KV-cache handoff between agents that combines token-level mixed-precision allocation, a self-contained CacheCard representation, and a HuggingFace-compatible cache injection path. Our current results support a narrower but clearer story than the original draft: on 150 GSM8K problems with Llama-3.1-8B-Instruct, adaptive quantization remains competitive under repeated handoff and shows its clearest gains against uniform quantization in deeper-hop, higher-budget settings; for handoff latency, the QKVShare path reduces TTFT relative to full re-prefill at every tested context, from 130.7 ms vs. 150.2 ms at nominal 1K context to 397.1 ms vs. 1029.7 ms at nominal 8K context;. Stage timing shows that post-injection generation, not card creation, dominates the current QKVShare latency path. These results position quantized KV handoff as a promising on-device systems direction while also highlighting the need for stronger controller ablations and apples-to-apples runtime comparisons.

# 1 Introduction

Edge-resident multi-agent LLM systems have a simple bottleneck: the sending agent already has the useful latent state, but the receiving agent usually reconstructs it from scratch. On-device deployments make this painful because KV caches grow linearly with context length and quickly dominate memory.

Current handoff choices are unsatisfying. Full re-prefill wastes latency, while FP16 KV sharing wastes memory. Quantization papers reduce single-agent KV footprint, and multi-agent sharing papers reduce redundant prefilling, but the combination is still lightly explored, especially for small-memory edge GPUs.

This paper therefore focuses on a narrower, evidence-backed question: how much can quantized KV handoff help on-device multi-agent inference today, and where do the remaining costs come from? We ground the story in measured multi-hop reasoning accuracy, TTFT after handoff, and resident-context density rather than in a broad end-to-end application sweep.

QKVShare packages a sender's KV state into a quantized CacheCard, uses a topology-aware controller to allocate bits when the adaptive path is enabled, and injects the received cache into the next agent. The current repository contains both the adaptive controller experiments (E1/E2) and a uniform-card handoff latency path (E3) used to isolate transport overhead.

Our main empirical takeaways are straightforward: quantized sharing avoids expensive re-prefill, and adaptive mixed-precision allocation is promising under repeated handoff. The results are encouraging, but the current evidence does not yet isolate a consistent topology-aware advantage over local-only adaptation, so this draft deliberately separates what is demonstrated from what remains open.

We make three concrete contributions:

Contribution 1: Cross-agent-aware token scoring. We extend a Don't Waste Bits style controller with downstream-demand and segment features so that bit allocation can account for both local fidelity and the next agent's needs.

Contribution 2: Quantized CacheCard handoff. We define a compact cache artifact for inter-agent transfer that packages quantized KV groups, bit-width assignments, and the metadata needed for receiver-side injection.

Contribution 3: Measured prototype analysis. We provide a working prototype and an end-to-end evaluation spanning multi-hop GSM8K accuracy and handoff TTFT, including a new stage-level breakdown showing where the current latency path spends time.

Together, these results argue for quantized KV handoff as a practical systems direction, even though the current prototype still relies on mixed runtimes and a HuggingFace-compatible dequantizing injection path.

# 2 Related Work

## 2.1 KV-Cache Quantization

KV-cache quantization has emerged as a primary technique for reducing the memory footprint of LLM inference. KIVI introduced per-channel key quantization and per-token value quantization, demonstrating that 2-bit KV cache is feasible with minimal accuracy degradation. KVQuant extended this to 10-million-token contexts via per-channel quantization with non-uniform datatypes. ZipCache proposed salient token identification to allocate higher precision to critical tokens. "Cache Me If You Must" (ICML'25) introduced adaptive quantization strategies that respond to token importance signals. Oaken (ISCA'25) integrated quantization into the hardware pipeline. Most recently, "Don't Waste Bits" (April 2026) proposed a learned controller that assigns per-token bit-widths from {2, 4, 8, FP16} using four features: frequency, quality score, attention variance, and entropy-based uncertainty. However, all these methods target **single-agent** inference and do not consider how quantized caches interact across agent boundaries.

## 2.2 Multi-Agent KV-Cache Sharing

The multi-agent KV sharing space has seen rapid progress in 2025–2026. KVCOMM (NeurIPS'25) identified the offset-variance problem when reusing KV caches across different prefix contexts and proposed an anchor-based framework that approximates cache offsets at runtime, achieving 70%+ reuse rates and 7.8× prefill speedup in 5-agent settings. Cache-to-Cache (C2C, ICLR'26) took a more ambitious approach, using learned neural projectors to fuse KV caches across *heterogeneous* models (different architectures and sizes), achieving 8.5–10.5% accuracy gains over individual models. LRAgent (February 2026) focused on the multi-LoRA setting, decomposing caches into a shared base component and low-rank adapter deltas. DroidSpeak (NSDI'26) enabled cache reuse across fine-tuned models by selectively recomputing critical layers. LatentMAS proposed training-free latent collaboration where agents share KV-cache working memory, achieving 83.7% token savings. Ramp Labs' Latent Briefing used attention-matching compaction to compress cross-agent context, achieving 49% token savings on LongBench v2.

Concurrent with our work, Q-KVComm (Kriuk, arXiv:2512.17914, Nov 2025) proposed adaptive layer-wise quantization for inter-agent KV cache transmission, achieving 5–6× compression. However, Q-KVComm differs from QKVShare in four key respects: (1) it allocates bit-widths at *layer* granularity (entire layers at 4-bit or 8-bit), whereas QKVShare operates at *per-token* granularity following the "Don't Waste Bits" insight that token importance varies within a layer; (2) it was evaluated only on 1.1–1.5B models, while our primary evaluation uses Llama-3.1-8B matching the standard multi-agent benchmark scale; (3) it does not target edge deployment or measure agent density under memory pressure; and (4) it lacks a topology-aware importance signal—its sensitivity profiling considers only local reconstruction error, not downstream agent demands.

Most prior multi-agent sharing systems still assume FP16/BF16 and server-class hardware. Q-KVComm begins to explore quantized inter-agent transfer, but at layer granularity and smaller model scales. We view QKVShare as part of this emerging design space, distinguished by token-level allocation, on-device density analysis, and an explicit attempt to incorporate cross-agent features into the controller.

## 2.3 On-Device LLM Inference

On-device LLM deployment has progressed rapidly. MobiLoRA (ACL'25) accelerates LoRA-based inference on mobile devices via context-aware KV cache optimization. Continuum addresses multi-turn agentic workload scheduling with KV cache time-to-live mechanisms. Qualcomm's QNN and Apple's MLX provide hardware-optimized inference runtimes, though neither natively supports cross-agent cache sharing. The Agent Memory system demonstrated that Q4 quantization enables 4× more agent contexts in fixed device memory, but its per-agent isolation means every agent handoff still requires either re-prefill or full decompression.

### 2.4 Agent Communication Protocols

Text-based inter-agent protocols—Google's Agent-to-Agent (A2A) and Anthropic's Model Context Protocol (MCP)—define standardized message formats for agent collaboration. These protocols operate at the text level: agents serialize their outputs to structured text and the receiving agent must re-process this text from scratch. C2C demonstrated that KV-cache-based communication is richer and faster than text-based exchange. Our work extends this insight to the edge setting, proposing a "cache card" format that could complement existing text-based protocols with latent state transfer.

## 3 Preliminaries and Problem Formulation

### 3.1 KV-Cache in Transformer Attention

In a decoder-only Transformer with L layers and H attention heads per layer, the KV cache for a sequence of n tokens stores key and value tensors $K^{(l,h)}$, $V^{(l,h)} \in \mathbb{R}^{n \times d}$ for each layer l and head h, where d is the head dimension. The total cache size scales as $O(2 \cdot L \cdot H \cdot n \cdot d \cdot b)$, where b is the bytes per element. For FP16, $b = 2$; for Q4 quantization, $b \approx 0.5$, yielding a 4× memory reduction.

### 3.2 Multi-Agent Cache Sharing Problem

We define a multi-agent system as a directed graph $G = (A, E)$ where $A = \{a_1, a_2, ..., a_k\}$ is a set of agents and $E \subseteq A \times A$ represents communication edges. When agent $a_i$ passes context to agent $a_j$ along edge $(a_i, a_j) \in E$, the receiving agent needs the sending agent's KV cache (or an equivalent representation) to avoid re-prefilling the shared context.

In the FP16 setting, this cache handoff is straightforward but memory-intensive. In the quantized setting, we face three challenges: (1) the quantization parameters (scales, zero-points) are computed for the sending agent's attention patterns and may not be optimal for the receiver, (2) positional encodings must be realigned when the receiver has a different prefix, and (3) quantization error accumulates across multiple hops in the agent graph.

### 3.3 Joint Optimization Objective

For each token t in the cache, we define two scoring functions:

**Local importance I(t):** Measures the token's contribution to the current agent's output quality. Following "Don't Waste Bits," this incorporates attention frequency, quality score, attention variance, and entropy-based uncertainty.

**Transferability T(t):** Measures the token's expected importance to downstream agents. This is derived from the agent topology: tokens in shared context segments, tokens with high cross-agent attention overlap (estimated from anchor pools), and tokens tagged as "shared state" by the agent protocol.

The combined importance score S(t) = α · I(t) + (1 − α) · T(t) determines the bit-width assignment b(t) ∈ {2, 4, 8, 16}, where α balances local accuracy against shareability. The optimization objective minimizes the expected quantization error subject to a memory budget M:

$$\min_e \sum_t \varepsilon(t, b(t)) \cdot S(t) \quad s.t. \quad \sum_t b(t) \cdot d \leq M$$

where ε(t, b(t)) is the quantization error for token t at bit-width b(t).

# 4 QKVShare: Method

QKVShare combines topology-aware bit allocation, a CacheCard handoff representation, and a prototype receiver-side injection path. The current implementation supports adaptive quantization for accuracy experiments and a HuggingFace-compatible dequantize-then-inject path for latency analysis; a fully fused quantized cross-agent attention kernel remains future work.

## 4.1 Agent-Topology-Aware Importance

The “Don’t Waste Bits” controller uses four per-token features—frequency, quality score, attention variance, and entropy—to assign bit-widths. We extend this controller with two additional features that encode cross-agent information:

### *4.1.1 Downstream Demand Signal*

Given the agent topology graph G, for each token t in agent $a_i$’s cache, we compute a downstream demand score D(t) that estimates how important t is for the next agent $a_j$ in the graph. We approximate this using two lightweight signals:

**Segment-level prior:** Tokens belonging to shared context segments (system prompts, retrieved documents, conversation history) receive a baseline demand boost, since these are by definition relevant to downstream agents. Segment boundaries are identified from the agent protocol metadata.

**Attention anchor overlap:** Building on KVCOMM’s insight that token importance is partially predictable from embedding similarity, we maintain an anchor pool of previously shared tokens and their downstream attention patterns. For a new token t, we find its nearest anchor and use the anchor’s downstream attention weight as a proxy for D(t). This is computed once during the first agent handoff and cached for subsequent hops.

### *4.1.2 Extended Controller*

The extended controller takes a 6-dimensional feature vector [frequency, quality, attention_variance, entropy, downstream_demand, segment_type] and outputs a bit-width assignment b(t) ∈ {2, 4, 8, 16}. The controller is a lightweight 2-layer MLP (512 parameters) trained via supervision on a held-out calibration set where ground-truth importance is computed by measuring the accuracy impact of quantizing each token individually.

## 4.2 Quantized Cache Handoff Protocol

We use the term CacheCard for the artifact passed between agents. In the current prototype it is optimized for in-memory transfer inside one runtime rather than for a finalized wire format.

#### *4.2.1 Cache Card Format*

A cache card C consists of:

Quantized KV groups: Keys and values are packed into adaptive or uniform quantized groups produced by the quantizer.

Metadata: The card tracks the token-level bit-width vector, sequence length, model and sender identifiers, and a position-offset placeholder for future cross-prefix alignment work.

Average-bit statistics: We record the achieved average bit budget so the receiver and the evaluation code can reason about compression ratio and memory footprint.

#### 4.2.2 Receiver-side Injection Path

For compatibility with HuggingFace Transformers, the current injection step reconstructs FP16 past_key_values before the receiving model performs its next forward pass. This lets us measure the practical cost of card creation, injection, and generation today while leaving a fully fused quantized attention kernel as future work.

### 4.3 Quantized Cross-Agent Attention Path

The long-term goal is to consume foreign quantized caches directly inside attention. The present repository does not yet expose a production fused Q4->Q4 kernel; instead, it uses the CacheCard abstraction to package quantized groups and a receiver-side injection path to study handoff overhead in a compatible stack.

This distinction matters for the latency results in Section 6.3: stage timing shows that model generation after injection dominates the measured TTFT, so the current bottleneck is the receiving forward pass rather than CacheCard construction itself.

A native fused kernel remains a clear optimization target and would make the adaptive handoff path more comparable to specialized runtime baselines.

## 5 Experimental Setup

### 5.1 Models and Devices

All adaptive experiments use Meta-Llama-3.1-8B-Instruct in a local PyTorch/HuggingFace stack with 4-bit weight quantization, while the baseline sharing and re-prefill measurements use llama.cpp on the same 12 GB NVIDIA RTX 5070 Ti Laptop GPU. The mixed-runtime setup is a limitation and is revisited in Section 7.

### 5.2 Agent Topologies

We evaluate two topology types:

We report quantitative results on sequential chains only: E1 spans 2-5 agent hops on GSM8K, E2 uses paired controller comparisons, E3 benchmarks a single handoff.

Hierarchical and heterogeneous pipelines motivated the design, but they are not yet part of the committed quantitative results and are therefore left out of the claims in this draft.

### 5.3 Tasks and Benchmarks

E1 evaluates 2-5 hop GSM8K chains on 150 problems with Llama-3.1-8B-Instruct. We report exact-match numeric accuracy from the final agent.

E3 measures TTFT after a single handoff using nominal 1K, 4K, and 8K synthetic contexts; the tokenizer counts in the JSON are approximately 476, 1939, and 3877 tokens respectively. The QKVShare TTFT row currently uses uniform Q4 CacheCards to isolate transport and injection cost.

### 5.4 Baselines

Different experiments require different baselines, so Table 1 summarizes the method families used throughout the paper.

| Method family | Where used in this draft |
|---|---|
| FP16 re-prefill | E3 lower-bound baseline: recompute the shared context at the receiver. |
| FP16 cache copy | E3 cache-restore baseline in llama.cpp with prompt-prefix reuse. |
| Uniform Q4 / Q8 share | E1 and E4 static quantization baselines; E3 uses a uniform Q4 CacheCard handoff path. |
| Adaptive local | E1/E2 local-only controller ablation at matched bit budgets. |
| Adaptive topology (QKVShare) | E1/E2 primary method: topology-aware mixed-precision allocation for cross-agent sharing. |

## 6 Results and Analysis

### 6.1 Quantization Error Propagation (E1)

E1 studies how accuracy changes when caches are handed across 2-5 agent hops. Table 2 reports the full 150-problem GSM8K results for FP16 sharing, uniform quantization, and the two adaptive controllers at matched approximately 4-bit and 8-bit budgets.

Table 2. GSM8K accuracy (%) across 2-5 agent hops (150 problems, Llama-3.1-8B-Instruct).

| Method | 2 hops | 3 hops | 4 hops | 5 hops |
|---|---|---|---|---|
| FP16 share | 81.33 | 76.67 | 75.33 | 72.67 |
| Uniform Q4 | 82.00 | 75.33 | 69.33 | 76.67 |
| Uniform Q8 | 82.67 | 73.33 | 69.33 | 71.33 |
| Adaptive local Q4 | 76.00 | 79.33 | 76.67 | 73.33 |
| QKVShare topo Q4 | 78.67 | 80.67 | 77.33 | 72.00 |
| Adaptive local Q8 | 76.67 | 80.00 | 82.00 | 83.33 |

| QKVShare topo Q8 | 76.00 | 79.33 | 82.00 | 82.67 |
|---|---|---|---|---|

Three patterns matter. First, uniform quantization becomes less stable as hop depth increases, especially at 4 and 5 hops. Second, the clearest adaptive signal in the current results is at approximately 8 bits: adaptive_topology_q8 reaches 82.67% at 5 hops versus 71.33% for uniform_q8, although it does not consistently beat the local-only adaptive controller. Third, the approximately 4-bit story is mixed rather than uniformly favorable for QKVShare. Because the adaptive rows are measured in a different runtime stack from the llama.cpp baselines, we treat E1 as directional evidence rather than a fully controlled head-to-head comparison.

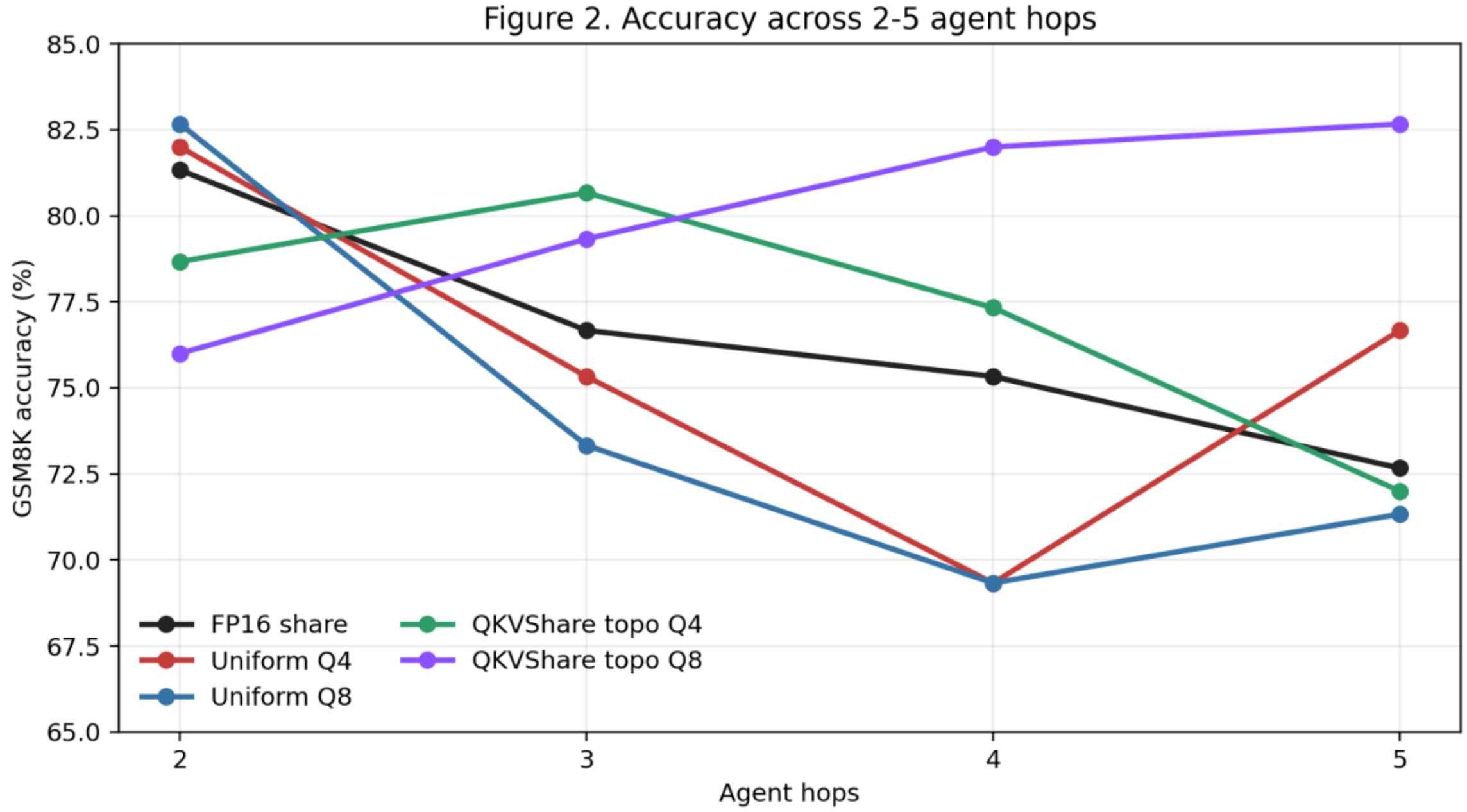

Figure 2. Representative E1 accuracy curves for FP16 sharing, uniform quantization, and topology-aware QKVShare across 2-5 hops.

## 6.2 Adaptive vs. Static Sharing (E2)

E2 isolates the controller question more directly. The committed E2 artifacts do not yet support the broad ablation originally promised, so we report the current paired-comparison evidence instead of overstating it.

At approximately 4 bits on 300 GSM8K problems, topology-aware allocation records 10 unique wins versus 8 for the local-only controller (net +2), and is essentially tied with uniform Q4 at 16 wins to 17 (net −1). Table 3 makes this explicit.

Table 3. Current paired E2 evidence at approximately 4 bits (300 GSM8K problems).

We therefore treat the topology-aware component as an open hypothesis rather than a settled win. The current repository suggests that adaptive quantization can help under repeated handoff, but it does not yet isolate a robust topology-over-local advantage at fixed budget.

| Comparison | Both correct | Both wrong | Topo only | Other only | Net |
|---|---|---|---|---|---|
| Topo Q4 vs local Q4 | 223 | 59 | 10 | 8 | +2 |

| Topo Q4 vs uniform Q4 | 217 | 50 | 16 | 17 | -1 |
|---|---|---|---|---|---|

### 6.3 Latency Benchmark (E3)

E3 benchmarks TTFT after handoff. Because the baselines run in llama.cpp while the QKVShare path runs through a HuggingFace-compatible PyTorch stack, we use the re-prefill comparison as the primary systems takeaway and treat native slot restore as a stronger specialized lower bound rather than as an apples-to-apples target.

Table 4. TTFT median (ms) after handoff at nominal 1K, 4K, and 8K shared context lengths (n = 2).

| Method | 1K | 4K | 8K |
|---|---|---|---|
| FP16 re-prefill | 150.2 | 565.3 | 1029.7 |
| FP16 cache copy | 21.7 | 26.1 | 26.5 |
| Q4 cache reload | 26.7 | 24.7 | 24.8 |
| QKVShare uniform Q4 handoff | 130.7 | 152.5 | 397.1 |

The QKVShare uniform-card handoff reduces TTFT relative to full re-prefill at every tested nominal context, from 130.7 ms vs. 150.2 ms at 1K to 397.1 ms vs. 1029.7 ms at 8K. Stage timing shows that generation dominates the current path: create-card / inject / generate medians are 4.4 / 13.8 / 112.4 ms at 1K, 21.0 / 17.8 / 113.7 ms at 4K, and 71.3 / 92.8 / 232.9 ms at 8K. This means the receiving forward pass, not CacheCard construction, is the largest remaining bottleneck.

### 6.4 What the Current Evidence Leaves Open

On the memory front, uniform Q4 already supports 2.8× more concurrent 8K contexts than FP16 on Llama-3.1-8B (64 vs 23 agents); a direct QKVShare density measurement requires a llama.cpp-backed implementation and is left for future work. The current repository does not yet include the broader HotpotQA, automotive, or thermal analyses described in early planning documents. For this draft, we therefore position QKVShare as a measured prototype paper centered on multi-hop reasoning accuracy, handoff TTFT, and memory density.

## 7 Discussion and Limitations

### 7.1 Error Accumulation

The multi-hop GSM8K results show that quantization error does accumulate, but not monotonically and not equally across budgets. The clearest robust benefit today appears in adaptive-versus-uniform comparisons at the higher 8-bit setting; the aggressive 4-bit regime remains noisy, and the added value of topology-aware features over local-only adaptation is still unresolved.

### 7.2 Mixed Runtime Stack

E1 and the llama.cpp portions of E3 use a different runtime from the PyTorch/HuggingFace QKVShare path. This lets us compare practical deployment options on the same machine, but it also means absolute latencies mix algorithmic differences with implementation-stack differences. A stronger apples-to-apples evaluation would run both paths in the same runtime or provide a native fused QKVShare kernel.

### 7.3 Prototype Handoff Path

The current handoff benchmark uses uniform Q4 CacheCards for latency isolation and reconstructs FP16 past_key_values during injection for HuggingFace compatibility. As a result, Section 6.3 measures the current prototype path rather than a fully adaptive end-to-end quantized attention kernel.

### 7.4 Hardware and Model Scope

All quantitative results use Llama-3.1-8B-Instruct on a single 12 GB NVIDIA RTX laptop GPU. We therefore avoid broad claims about mobile NPUs, heterogeneous models, or cross-hardware generality in this draft.

### 7.5 Highest-Value Next Experiment

If we add one experiment before treating the paper as stronger than an internal/arXiv-ready draft, it should be a rerun of E2 with full-scale paired statistics at both approximately 4 and 8 average bits. That experiment would directly test the paper's most reviewer-sensitive claim: whether topology-aware allocation consistently outperforms local-only or uniform allocation at fixed budget.

## 8 Conclusion

QKVShare explores quantized KV-cache handoff as a systems primitive for on-device multi-agent inference. The current evidence supports two concrete points: avoiding re-prefill materially improves TTFT, and adaptive mixed-precision allocation is promising under repeated handoff even though the specific topology-aware contribution is not yet conclusively isolated.

Just as importantly, the current prototype clarifies what remains to be done. The E3 stage breakdown shows that the post-injection model forward dominates latency, and the current E2 evidence is not yet strong enough to support a sweeping topology-aware claim at aggressive 4-bit budgets. We therefore view this draft as a strong measured foundation for quantized latent handoff, with a clear next step: a fully matched adaptive handoff path and a stronger controller ablation.

**AI Assistance**

The author used Cursor (an AI-assisted IDE) and Claude Code (Anthropic's CLI) as coding and writing assistants during the development of experiments, analysis scripts, and manuscript drafting. All experimental design, results interpretation, and scientific claims are the author's own.